\documentclass[lettersize,journal]{IEEEtran}
\usepackage{amsmath,amsfonts}
\usepackage{algorithmic}
\usepackage{algorithm}
\usepackage{array}
\usepackage[caption=false,font=normalsize,labelfont=sf,textfont=sf]{subfig}
\usepackage{textcomp}
\usepackage{stfloats}
\usepackage{url}
\usepackage{verbatim}
\usepackage{graphicx}
\usepackage{cite}

\usepackage{threeparttable}
\usepackage{times}
\usepackage{epsfig}
\usepackage{graphicx}
\usepackage{amsmath}
\usepackage{amssymb}
\usepackage{helvet} 
\usepackage{courier}  
\usepackage{graphicx}  
\usepackage{epsfig}
\usepackage{amsmath}
\usepackage{amssymb}
\usepackage{bm}
\usepackage{subfloat}
\usepackage{booktabs}
\usepackage{multirow}
\usepackage{float}
\usepackage{bbding}
\usepackage{array}
\usepackage{color}
\usepackage{diagbox}
\usepackage{comment}

\usepackage{algorithm}

\begin{document}

\title{Exploring Effective Mask Sampling Modeling for Neural Image Compression}

\author{Lin Liu, Mingming Zhao, Shanxin Yuan, Wenlong Lyu, Wengang Zhou \\Houqiang Li~\IEEEmembership{Fellow,~IEEE,} Yanfeng Wang, Qi Tian~\IEEEmembership{Fellow,~IEEE}

\thanks{Corresponding author: Mingming Zhao}}

\markboth{}%
{Shell \MakeLowercase{\textit{et al.}}: A Sample Article Using IEEEtran.cls for IEEE Journals}


\maketitle

\begin{abstract}
Image compression aims to reduce the information redundancy in images.
Most existing neural image compression methods rely on side information from hyperprior or context models to eliminate spatial redundancy, but rarely address the channel redundancy. Inspired by the mask sampling modeling in recent self-supervised learning methods for natural language processing and high-level vision, we propose a novel pre-training strategy for neutral image compression.
Specifically, Cube Mask Sampling Module (CMSM) is proposed to apply both spatial and channel mask sampling modeling to image compression in the pre-training stage.  
Moreover, to further reduce channel redundancy, we propose the Learnable Channel Mask Module (LCMM) and the Learnable Channel Completion Module (LCCM). Our plug-and-play CMSM, LCMM, LCCM modules can apply to both CNN-based and Transformer-based architectures, significantly reduce the computational cost, and improve the quality of images. 
Experiments on the public Kodak and Tecnick datasets demonstrate that our method achieves competitive performance with lower computational complexity compared to state-of-the-art image compression methods.

\end{abstract}

\section{Introduction}
\label{sec:intro}


Lossy image compression is a critical and challenging topic in both academia and industry, as it facilitates image storage and transmission.
Recently, neural image compression has demonstrated its powerful ability in this field, surpassing traditional hand-crafted image compression methods~\cite{GregoryKWallace1992TheJS,MajidRabbani2002AnOO,GaryJSullivan2012OverviewOT,ohm2018versatile}.
A typical neural image compression model comprises an encoder, a decoder, and an entropy model.
The encoder and the decoder work in tandem to enhance the non-linear representation ability of the network, enabling the model to reduce image redundancy effectively.

Image information redundancy includes spatial redundancy and channel redundancy. Even though spatial redundancy has been widely addressed, channel redundancy has received less attention in the literature. Most existing neural image compression methods rely on side information from hyperprior~\cite{JohannesBall2018VariationalIC,DavidMinnen2018JointAA,ZhengxueCheng2020LearnedIC} or context models~\cite{DavidMinnen2018JointAA,JooyoungLee2018ContextadaptiveEM,AKoyuncu2022CONTEXTFORMERAT,He_2021_CVPR} to eliminate spatial redundancy. 
%
%
These models capture the structure of edges and textured regions, and the differences between visually similar patches. 
However, the gap between training and testing data may lead to inadequate correlation of similar features for testing data, resulting in more consumption of code rate and unsatisfactory performance.

%
%
%

%
%
%
%

 \begin{figure}[t]
 		\centering  
  \includegraphics[width=0.48\textwidth]{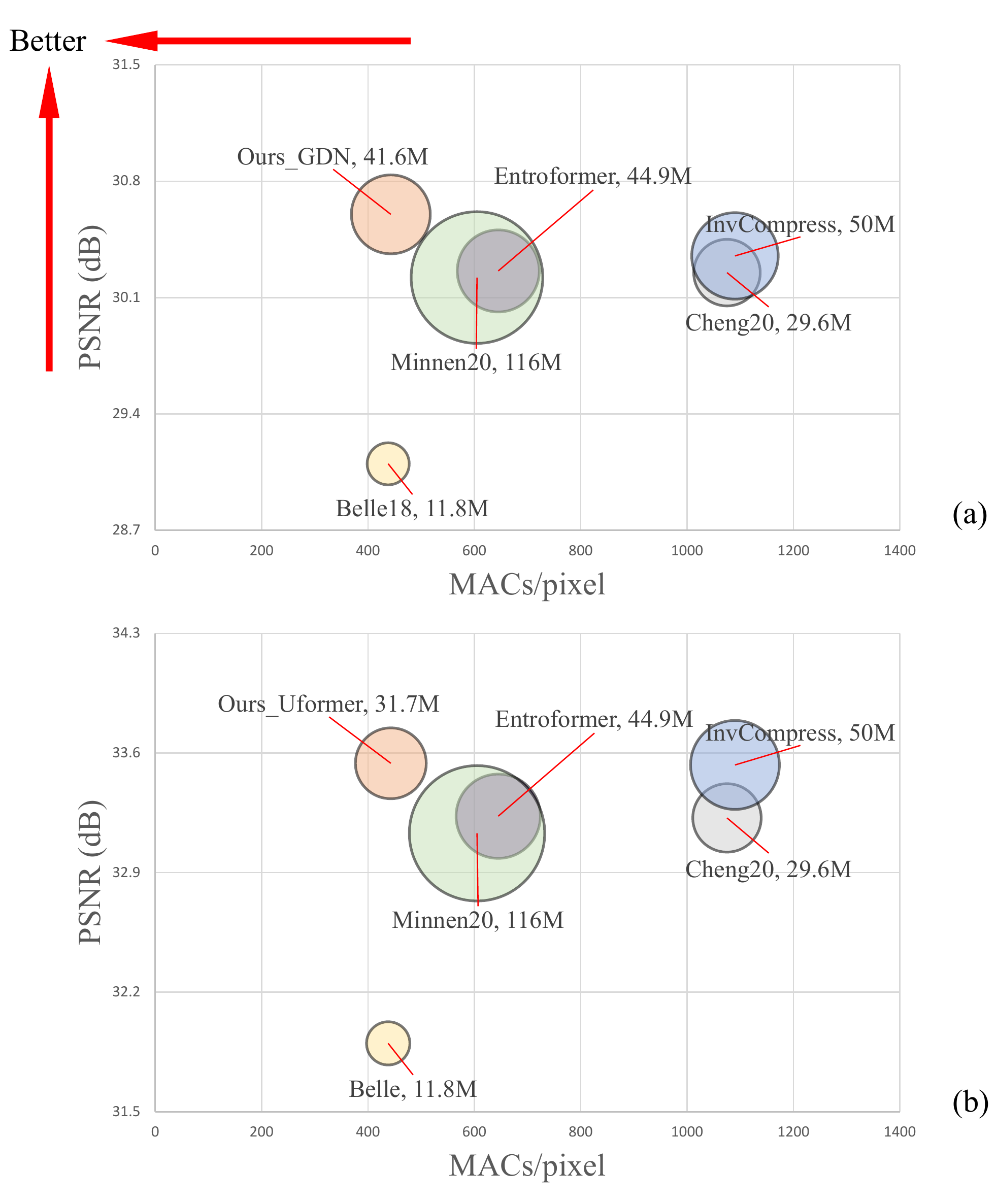}
 		\caption{Distortion, model size, and computation complexity comparisons between some state-of-the-art neural image compression methods and ours (Ours\_GDN) on the Kodak dataset when the bit-per-pixel (bpp) is 0.2 Figure(a) and the bit-per-pixel (bpp) is 0.4 Figure(b), respectively. The circle's size represents the corresponding network's size. The closer the circle is to the upper left corner of the plot, the better the corresponding method.} 
 		\label{fig:index}
 \end{figure}

To address channel redundancy, Jia et al.~\cite{JiaxiongQiu2021SlimConvRC} demonstrate that the deep features in neural networks are not compact and there exists redundancy among them. Channel pruning techniques \cite{YihuiHe2017ChannelPF,ZhuangweiZhuang2018DiscriminationawareCP,YuZhao2022DifferentiableCS} are used in Neural Architecture Search (NAS) that aims to compress layers of the network, but these methods are mainly used for high-level vision tasks, such as recognition~\cite{KaimingHe2016DeepRL}. 
However, channel redundancy has been less discussed in image compression. Han et al.~\cite{ChaoyiHan2020TowardVG} propose a method for variable-rate compression by masking positions where channels have strong correlations and only preserving the channel-wise mean value, but this method is specific for variable-rate image compression and not competitive with the most advanced image compression methods. 

%
%
%
%
%

In this paper, we draw inspiration from recent self-supervised learning methods based on mask image modeling~\cite{ZhendaXie2021SimMIMAS,KaimingHe2021MaskedAA,ChenWei2023MaskedFP,HangboBao2021BEiTBP,ChristophFeichtenhofer2023MaskedAA} and explore the application of mask sampling modeling in image compression.
We introduce a pre-training stage that includes a module to randomly mask some elements of the deep feature, enabling the masked element to gather information from other elements.
This operation can enhance the correlation between different elements in both spatial and channel domains.
We call this module the Cube Mask
Sampling Module (CMSM), which applies both spatial and channel mask sampling modeling. 
Furthermore, we propose the Learnable Channel Mask Module (LCMM) and Learnable Channel Completion Module (LCCM) to explicitly reduce channel redundancy.
For low-level vision tasks, some channels may not express information while others may express image-independent prior information.
%
These two kinds of channels can be eliminated through learnable channel masks,
which can reduce computation complexity while maintaining compression performance.
This explicit channel reduction operation can effectively reduce channel redundancy.

%
%
%
%
%
%
%

The experiments on the Kodak dataset and the Tecnick dataset demonstrate that our proposed methods achieve the competitive performance compared with state-of-the-art image compression methods. Our method performs particularly better at lower bit rates. 
Specifically, on the Kodak dataset, our model achieves state-of-the-art performance by reaching about 14\% rate savings for about 28dB anchored with VTM 16.2 (Fig.~\ref{fig:ratesaveing_results}).
In Fig.~\ref{fig:index}(a), we show the distortion, model size, and computation complexity comparison between state-of-the-art neural image compression methods and ours.
Our method achieves superior performance with less computation.
In addition, our method is suitable for both CNN-based and Transformer-based encoding/decoding structures.
The contributions of our work are as follows:

\begin{itemize}
  \item We explore effective mask sampling modeling for neural image compression. Specifically, we propose the Cube Mask Sampling Module (CMSM) for pre-training, the Learnable Channel Mask Module (LCMM) and Learnable Channel Completion Module (LCCM) to explicitly reduce channel redundancy.  
  
  \item Extensive experiments on both Kodak and Tecnick datasets show that our method achieves competitive performance with state-of-the-art image compression methods.

   \item Our plug-and-play method can be used in both CNN-based and Transformer-based architectures and achieves superior performance with less computation.
  
\end{itemize}

\section{Related Work}
\label{sec:formatting}
\subsection{Lossy Image Compression}
Traditional image compression methods are used in most image compression standards, such as the JPEG~\cite{GregoryKWallace1992TheJS}, JPEG2000~\cite{MajidRabbani2002AnOO}, HEVC~\cite{GaryJSullivan2012OverviewOT}, and VVC~\cite{ohm2018versatile} standards. These methods usually consist of the intra prediction module, frequency transform module, quantization module, and entropy coder. They try to achieve better rate-distortion optimization performance through carefully hand-crafted designs for one or more modules. 
These separate optimization methods are sub-optimal and often need manual parameter adjustment.

In the deep learning era, neural networks have been adopted for image compression, these methods are named as neural image compression. Belle et al.~\cite{JohannesBall2017EndtoendOI} have made pioneering work in this field, where they use a neural network to construct an image compression pipeline including an encoder, a decoder, and an entropy model. This pipeline is later widely adopted. 
Balle et al.~\cite{JohannesBall2018VariationalIC} later propose a hyperprior to capture spatial dependencies among the elements in the latent code, and thus can make the performance better by reducing spatial redundancy.
To further reduce spatial redundancy, more accurate estimation of distributions of latent codes have been explored, including mean and scale Gaussian distribution~\cite{DavidMinnen2018JointAA}, context model~\cite{DavidMinnen2018JointAA,JooyoungLee2018ContextadaptiveEM}, discretized Gaussian Mixture Likelihoods~\cite{ZhengxueCheng2020LearnedIC}, and transformer-based entropy model~\cite{JunHyukKim2022JointGA,YichenQian2022EntroformerAT}.

Another line of research~\cite{WeiguiLi2020DeepIC,ZhengxueCheng2019DeepRL,YueqiXie2021EnhancedIE,GeGao2021NeuralIC,DiptiMishra2021WaveletBasedDA,ChajinShin2022ExpandedAS,ZhisenTang2022JointGA} focus on designing better encoders and decoders, which are effective for modeling nonlinear transforms and obtain better performance. For example, Xie et al.~\cite{YueqiXie2021EnhancedIE} introduce an invertible convolution block for the projection between images and latent codes. 
%
Zou et al.~\cite{Zou_2022_CVPR} propose a new Transformer-based encoder/decoder which can make full use of both global structure and local texture.
These specially designed encoders/decoders are heavily intertwined with the chosen backbone architectures. We propose plug-and-play modules that can be embedded with any type of backbone architecture, and demonstrate its use in both CNN-based and Transformer-based models. Specifically, we propose mask sampling modules that are placed between the encoder and the entropy model to enhance the nonlinear representation ability.
%
Our method is different with some channel-wise image compression methods such as Contextformer~\cite{AKoyuncu2022CONTEXTFORMERAT} and Minnen20~\cite{DavidMinnen2020ChannelwiseAE}. \cite{AKoyuncu2022CONTEXTFORMERAT} incorporates both spatial and channel attention mechanisms within a context model. The approach introduced in \cite{AKoyuncu2022CONTEXTFORMERAT} divides the encoded features $y$ into several different parts based on their channels, and then processing them with different branches. Their method of handling channels does not involve dropping or replenishing channels.
Compared to others, our method does not require meticulous network design and is more adaptable to different networks.

\subsection{Mask Sampling Modeling}

Masked sampling modeling holds out a portion of the input elements and train models to predict the missing information.
It has been widely used in both natural language processing \cite{JacobDevlin2018BERTPO,YinhanLiu2019RoBERTaAR} and computer vision \cite{ZhendaXie2021SimMIMAS,KaimingHe2021MaskedAA,ChenWei2023MaskedFP,HangboBao2021BEiTBP,ChristophFeichtenhofer2023MaskedAA}, and proves powerful and effective.
Masked language modeling~\cite{JacobDevlin2018BERTPO,YinhanLiu2019RoBERTaAR} is an effective and powerful self-supervised learning approach in natural language processing (NLP), and achieves good performance. For example, the pre-trained BERT~\cite{JacobDevlin2018BERTPO} model can be finetuned and achieves new state-of-the-art results on eleven NLP tasks.
In computer vision, the performance of the recent self-supervised learning methods based on mask image modeling have achieved convincing performance on high-level vision tasks, such as classification, detection, and semantic segmentation~\cite{ZhendaXie2021SimMIMAS,KaimingHe2021MaskedAA,ChenWei2023MaskedFP,HangboBao2021BEiTBP,ChristophFeichtenhofer2023MaskedAA}. They are gradually outperforming the competitive of the self-supervised learning algorithm based on pretext tasks and contrastive learning~\cite{SpyrosGidaris2018UnsupervisedRL,he2019moco,chen2020mocov2}. 

Until recently, mask sampling modeling has rarely been discussed or researched in low-level vision tasks such as image generation \cite{chang2022maskgit}, image restoration \cite{YuhongzeZhou2021ViewBA}, and image compression \cite{AlaaeldinElNouby2022ImageCW}. 
Chang et al.~\cite{chang2022maskgit} propose MaskGIT for image generation, where the model predicts randomly masked tokens during training.
Zhou et al.~\cite{YuhongzeZhou2021ViewBA} propose a plug-and-play Mask Guided Residual Convolution for blind-spot based denoising.
Very recently, El-Nouby et al.~\cite{AlaaeldinElNouby2022ImageCW} utilize mask image modeling method for conditional entropy model in image compression. Specifically, they partition the image features into subsets and sequentially predict the image patch indices in several stages.
This method mainly focuses on improving the entropy model itself, while ours adds a module between the encoder and the entropy model.


\section{Method}
\label{sec:Method}

Based on quantization types, there are two groups of image compression methods: scalar quantization-based methods~\cite{JohannesBall2017EndtoendOI,JohannesBall2018VariationalIC,DavidMinnen2018JointAA,JunHyukKim2022JointGA,YichenQian2022EntroformerAT,DavidMinnen2018JointAA} and vector quantization-based methods~\cite{AlaaeldinElNouby2022ImageCW,XiaosuZhu2022UnifiedMG}. Our method falls into the category of the scalar quantization-based methods, which are more widely used and usually achieve better performance within the broader bit-per-pixel (bpp) range. We first briefly introduce the common components of scalar quantization-based methods in Sec 3.1, and then describe our method in Sec 3.2, 3.3, and 3.4.

\subsection{Background}

 \begin{figure*}[t]
 		\centering
 		\includegraphics[width=0.95\textwidth]{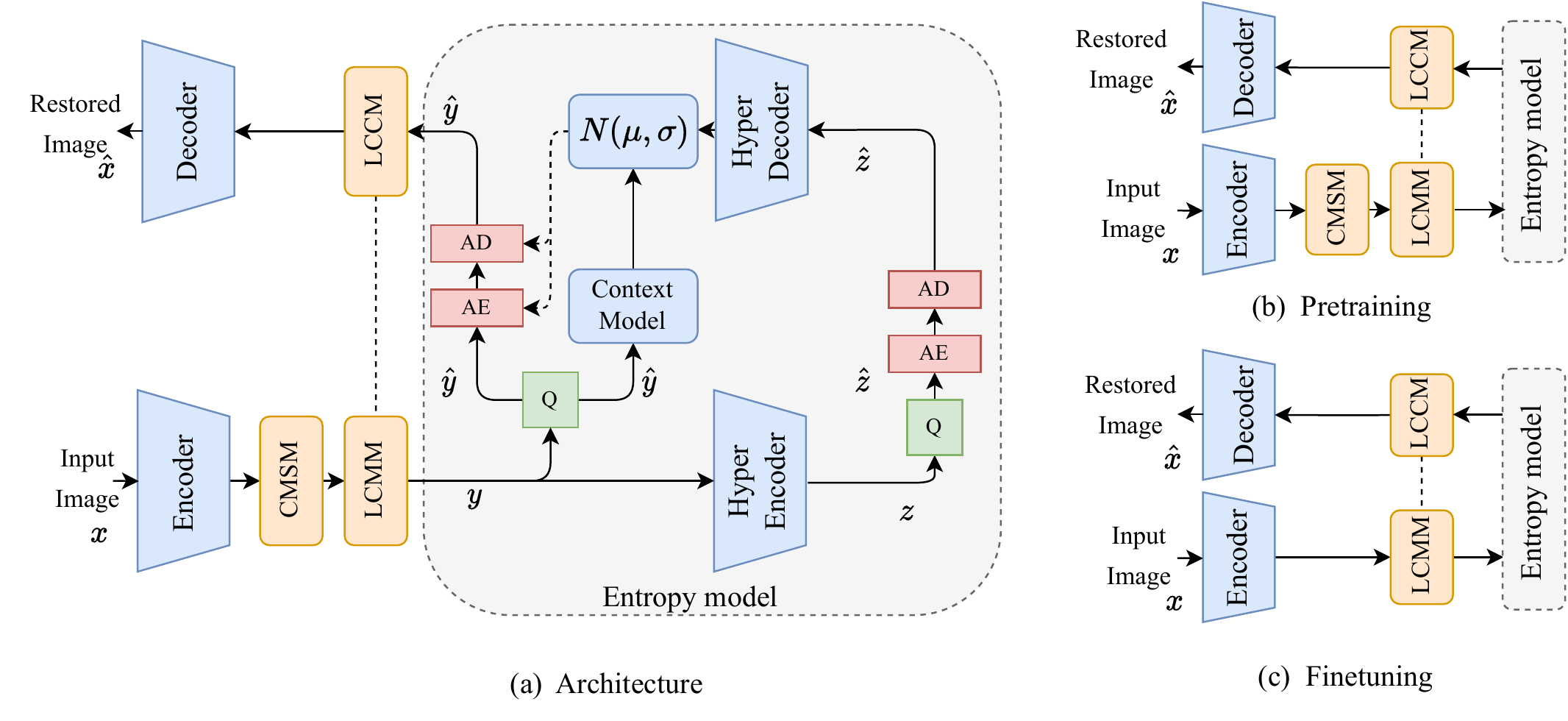}
 		\caption{(a) The architecture of our model, which contains encoder, decoder, entropy model, Cube Mask Sampling Module (CMSM), Learnable Channel Mask Module (LCMM), and Learnable Channel Completion Module (LCCM). (b) and (c) are the illustration of pretraining and finetuning.}
 		\label{fig:archtecture}
 \end{figure*}
 
\textbf{Loss Function for Image Compression.} An image compression method compresses an image $x$ into a bit stream and then restore a predicted image $\hat{x}$. The goal for image compression is to achieve a short bit stream and restore the original input image to the maximum extent, which can be formulated by,

\begin{equation}
\label{eq:lossfunction}
\underbrace{\mathbb{E}_{\boldsymbol{x} \sim p_{\boldsymbol{x}}}\left[-\log _2 p_{\hat{\boldsymbol{y}}}\left(\left\lfloor f_a(\boldsymbol{x})\right\rceil\right)\right]}_{\text {rate }}+\lambda \cdot \underbrace{\mathbb{E}_{\boldsymbol{x} \sim p_{\boldsymbol{x}}}\left[d\left(\boldsymbol{x}, \boldsymbol{\hat{x}}\right)\right]}_{\text {distortion }},
\end{equation}

\begin{equation}
\hat{\boldsymbol{y}} = \lfloor f_a(\boldsymbol{x})\rceil, \ \boldsymbol{\hat{x}} = f_s\left(\left\lfloor f_a(\boldsymbol{x})\right\rceil\right),
\end{equation}
where $\lambda$ is a hyper-parameter for controlling the trade-off between the bit stream's rate and the distortion between $\hat{x}$ and $x$; $f_{a}(\cdot)$, $\lfloor\cdot\rceil$, and $f_{s}(\cdot)$ represent the encoder, the quantizer, and the decoder, respectively; $\hat{y}$ represents the quantized latent features. $p_{x}$ is the distribution of input source image and $p_{\hat{y}}$ is the probability distribution of $\hat{y}$.
The distortion $d$ is defined as $MSE(x, \hat{x})$ for MSE-oriented model
and $1 - MS\text{-}SSIM(x, \hat{x})$ for MS-SSIM -oriented model.

\textbf{Hyper-prior and Context Model.}
Hyperprior is a type of prior that guides the selection of an appropriate entropy model (traditional methods) or the determination of its parameters (deep-learning-based methods).
Hyper-prior model is effective to parameterize the distributions of the quantized latent features $\hat{y}$ and achieve good performance. 
Specifically, (see the entropy model of Fig.~\ref{fig:archtecture}(a)) the latent representation $y$ is fed into the hyper encoder $h_{a}$, which then generates a set of coding variables $z$.
%
By modeling $y$ as a vector of zero-mean Gaussian distribution with standard deviation $\sigma$, we can obtain the quantized variable $\hat{z}$, which is then fed into the hyper decoder $h_{s}$. The output of $h_{s}$ provides an estimation of certain parameters of the distribution of $\hat{y}$.
For example, Belle~\cite{JohannesBall2017EndtoendOI} estimates standard deviations $\hat{\sigma}$.
Minnen~\cite{DavidMinnen2018JointAA} estimates both mean $\hat{\mu}$ and deviation $\hat{\sigma}$. Thus, the first term in Eqn.~\eqref{eq:lossfunction} can be extended to two terms,
\begin{equation}
\mathbb{E}\left[-\log _2 p_{\hat{\mathbf{y}} \mid \hat{\mathbf{z}}}(\hat{\mathbf{y}} \mid \hat{\mathbf{z}})\right]+\mathbb{E}\left[-\log _2 p_{\hat{\mathbf{z}} \mid \theta}(\hat{\mathbf{z}} \mid \theta)\right].
\end{equation}

In ~\cite{DavidMinnen2018JointAA}, Minnen et al. propose a context model, which automatically estimates the distribution parameters of current latent element $\hat{y_{i}}$ using the previously estimated elements $\hat{y}_{<i}$.
Qian et al.~\cite{YichenQian2022EntroformerAT} propose Entroformer which has a transformer-based context model. Because of its decoding speed and  competitive performance, we use this transformer-based context model in our method.

\subsection{Architecture and Training Process}
Fig.~\ref{fig:archtecture}(a) shows the architecture of our method. In addition to the standard modules of an encoder, a decoder, an entropy model, our architecture also have the novel Cube Mask Sampling Module (CMSM),
Learnable Channel Mask Module (LCMM), and Learnable Channel Completion Module (LCCM). 
After the input image is encoded into a feature by the encoder, this feature is sequentially passed through our designed CMSM and LCMM. After that, we obtain $y$ and send it into the entropy model, and get the $\hat{y}$ from the entropy model.
The CMSM and LCMM modules mask out spatial and channel features, while LCCM completes the missing channel of $y$, and finally, we can obtain the recovered image $\hat{x}$ decoded by the decoder.

As shown in Fig.~\ref{fig:archtecture}(b) and (c), our training pipeline includes two stages: pretraining and finetuning. All modules are used in the pretraining stage, while the CMSM module is removed during the finetuning stage. Since the CMSM module does not modify the size the features, when removed in the finetuning stage, it does not affect the flow of the pipeline.


 \begin{figure*}[t]
 		\centering		\includegraphics[width=0.95\textwidth]{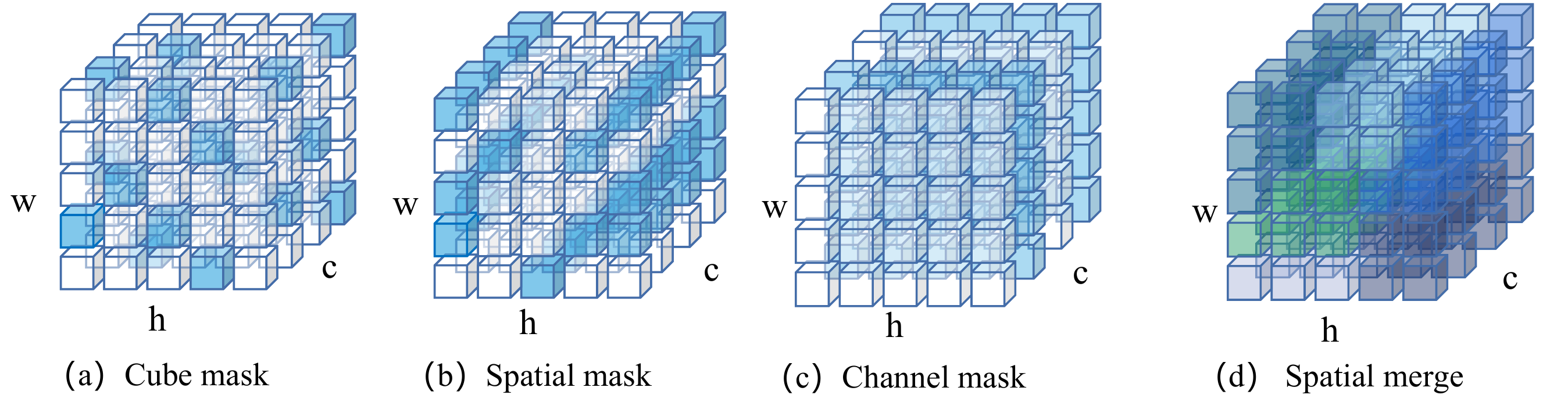}
 		\caption{The illustration of different kinds of mask sampling/ merging methods.}
 		\label{fig:maskmethod}
 \end{figure*}
 
\subsection{Cube Mask Sampling Module}
Through the encoder, input image $x$ turns into the deep feature $F \in R^{h \times w \times c}$, where $h$ and $w$ denote the height and width of the feature maps, $c$ is the number of channels. $F$ is then sent into the CMSM module.
In the CMSM module, we randomly select some elements of $F$ and set their values to 0. 
This random sampling strategy, named `cube mask', is agnostic to the spatial dimension and channel dimension of the feature.
As shown in Fig.~\ref{fig:maskmethod}(a), in cube mask, the blue elements are the original elements, while the white elements are the masked elements and their values are set to 0.

This random mask strategy is more effective than structured mask strategies such as spatial mask and channel mask (see Fig.~\ref{fig:maskmethod} (b) and (c)), we demonstrate it through experimental comparisons in the experiment section). The pre-training strategy of spatial (or channel) mask makes it difficult to obtain information from other spatial (or channel) positions in the masked space (or channel) when the masked information is special, such as complex textures or contains high-frequency information.
We also compare with the spatial merge strategy in~\cite{CedricRenggli2022LearningTM}. As shown in Fig.~\ref{fig:maskmethod} (d), this strategy replace several spatially similar elements with their mean values. This spatial merge strategy is very effective on high-level vision tasks~\cite{CedricRenggli2022LearningTM}, but is not good on low-level vision tasks, where the average means that features are filtered by low-pass filters. 
It is usually difficult to recover high-frequency information only from low-frequency information~\cite{RunyuanCai2021FreqNetAF}.

When adopting the random mask strategy, the optimal masking ratio is a key parameter to tune as it depends on the redundancy of the data used~\cite{KaimingHe2021MaskedAA}.
For example, BERT~\cite{JacobDevlin2018BERTPO} uses the rate of 15\% for the language masking, while MAE~\cite{KaimingHe2021MaskedAA} uses 75\% for the image masking, which indicates that there is more redundancy information in images than languages. However, MAE deals with high-level vision tasks, while we are dealing with low-level vision tasks, our optimal masking ratio should be lower than that of MAE.
%
%
Moreover, the optimal masking ratio also varies along with the constraints of different compressed bit streams. The compression with higher bit stream is smaller, so the redundancy of information is lower, and the optimal masking ratio is lower. 

 \begin{figure}[t]
 		\centering		\includegraphics[width=0.45\textwidth]{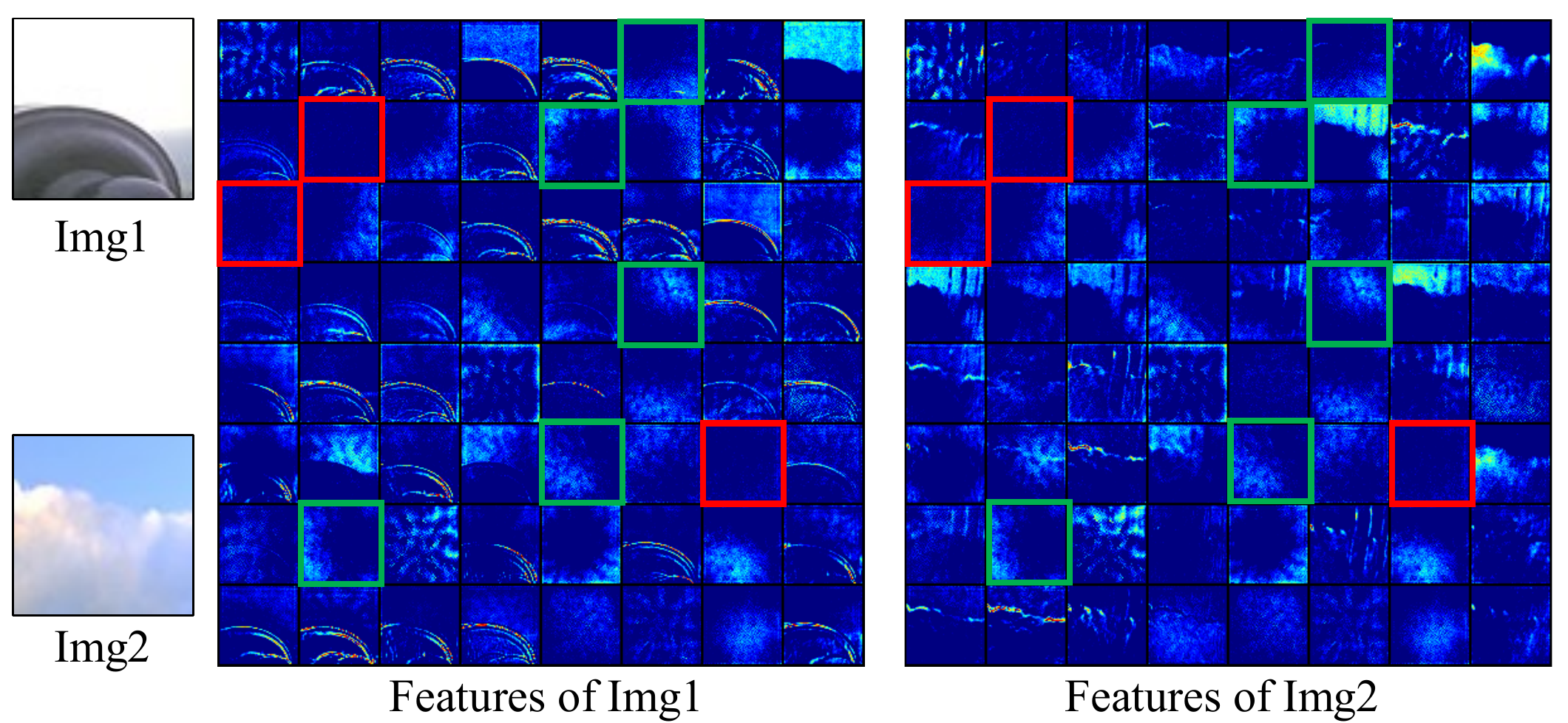}
 		\caption{Visualization of all 64 channels of
the output features of the pre-trained PLM module in \cite{LinLiu2022TAPETP}. Red boxes: nearly empty contents. Green boxes: image-independent spatial association prior information.}
 		\label{fig:vischannel}
 \end{figure}

  \begin{figure}[t]
 		\centering		\includegraphics[width=0.40\textwidth]{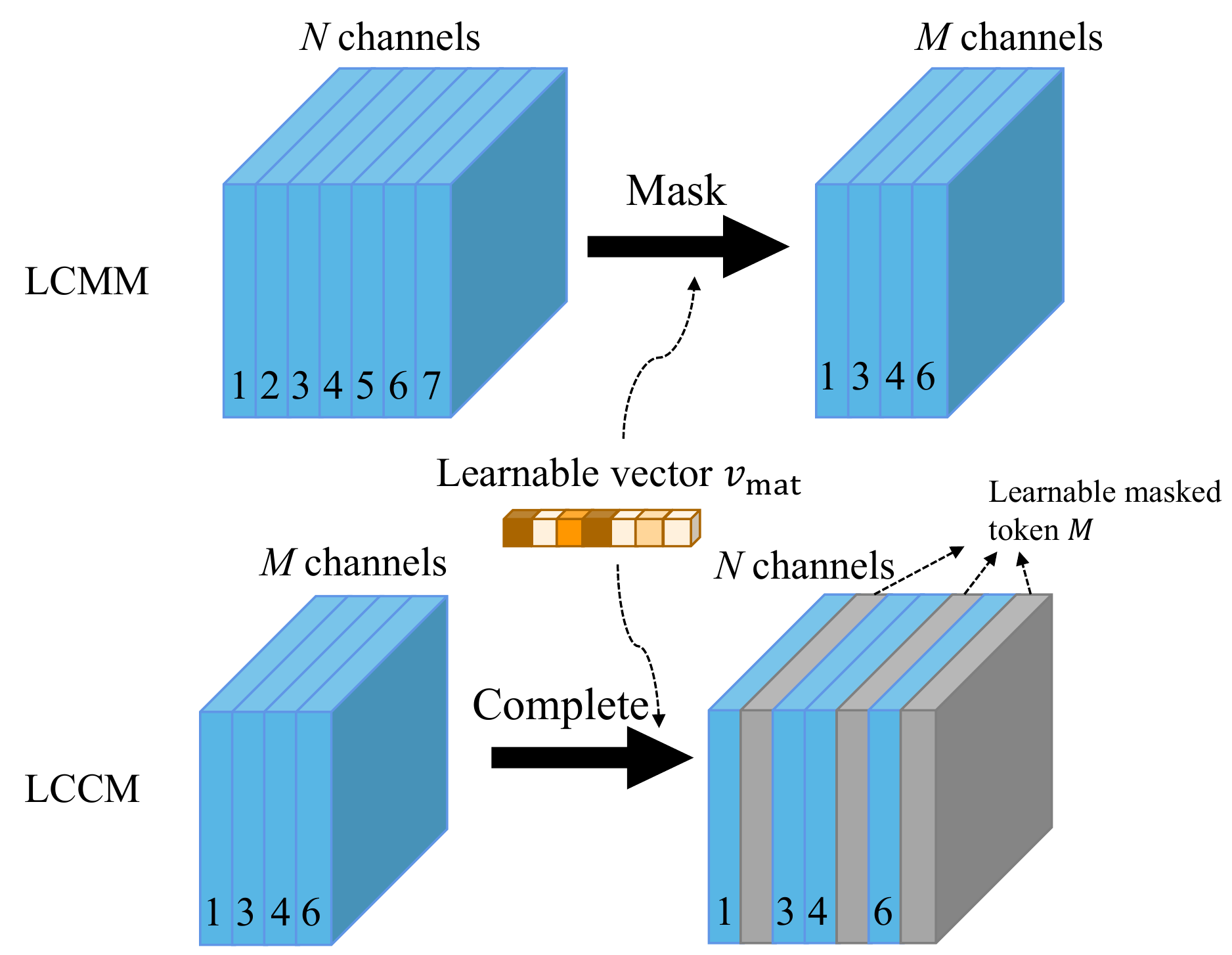}
 		\caption{The illustration of LCMM and LCCM .}
 		\label{fig:lcmm}
 \end{figure}
 
\subsection{Learnable Channel Mask Module}
The contents vary significantly among different channels of the feature maps. Some channels have little (or none) information~\cite{JiaxiongQiu2021SlimConvRC,JinhuaLiang2020ChannelCR,ShaohuiLin2017ESPACEAC}, and some channels express image-independent prior information~\cite{LinLiu2022TAPETP}. We visualize some channels of the output feature of the pre-trained PLM module of a recent paper for the low-level multi-task training~\cite{LinLiu2022TAPETP}. As shown in Fig.~\ref{fig:vischannel}, some feature maps marked with red boxes have almost no content; Some feature maps marked with green boxes show the spatial information, which shows almost the same patterns in different images.
Sending all the channels into the entropy model will lead to redundancy of information and increase of computation cost.
Therefore, we propose Learnable Channel Mask Module and Learnable Channel Completion Module to further reduce the redundant channels.

Assume that the dimension of the deep feature input to the LCMM module is $H \times W \times C$ and $C=\{c_{0},c_{1},c_{2},...,c_{N}\}$. A learnable vector $v_{mat} \in \mathbb{R}^{1 \times N}$ indicates the probability of each corresponding channel being selected. Then we select the $M$ channels with the highest probability and concat them as the input of the entropy model. 

For Learnable Channel Completion Module, the input $M$ channels are placed to its original place according to $v_{mat}$ and other channels are embedded with learnable mask tokens. Finally, LCCM outputs the feature with the size $H \times W \times C$. Fig.~\ref{fig:lcmm} shows the process of LCMM and LCCM.

\section{Experiments}
In this section, we conduct extensive experiments to evaluate the effectiveness and efficiency of our proposed method. Firstly, we show the implementation details and setup of our paper. Then we show the Ratio-Distortion (R-D) performance comparisons and visual comparisons with other methods. Finally, we show the ablation study.
%

\subsection{Experimental Setup}

\textbf{Network Implementation.}
For the encoder/decoder, we use the Uformer~\cite{ZhendongWang2022UformerAG} (with two basic Uformer blocks) when the BPP is lower than 0.4, and the Generalized
Divisive Normalization (GDN)~\cite{JohannesBall2017EndtoendOI} (with four basic blocks) when the BPP is bigger than 0.4.
The entropy model in~\cite{YichenQian2022EntroformerAT} is introduced, which contains 6 transformer-based hyper encoder blocks and 6 hyper decoder blocks.
%
More network details can be seen in the supplementary material.

\textbf{Training Details.} We use ImageNet-Val dataset~\cite{OlgaRussakovsky2014ImageNetLS} as our training dataset. This dataset is composed of 50,000 natural images.
To make the model adapt to different image resolution, we randomly downsample the training images as data augmentation. We randomly crop 256×256 image patches and set the batch size to 8. We use the AdamW optimizer~\cite{IlyaLoshchilov2019DecoupledWD}
The training procedure includes two stages: 1) Pretraining: In this stage, both Cube Mask Module and Channel Mask Modules are used. 2) Finetuning: only Channel Mask Modules are used.

In pretraining, we train our model for 300 epochs. The learning rate is initialized to be $1 \times 10^{-4}$ and decrease to 0.8 times of the former one in the 100 and 200 epoch. 
Our models are optimized with the rate-distortion trade-off loss function in \ref{eq:lossfunction}. We train our models with $\lambda$ equals to 0.3 and 600 when using MSE loss and MS-SSIM loss, respectively.

In finetuning, we remove the Cube Mask Module and train other networks for 1000 epochs. 
We set the initial learning rate to $8 \times 10^{-5}$, which is turned down to half of the former one in the 200, 400, and 800 epochs.

\textbf{Evaluation.} We evaluate our method on Kodak dataset \cite{Kodak1993} and the Tecnick dataset \cite{tecnick2014}.
The Kodak dataset has 24 natural images, each with the 768×512 resolution. The Tecnick dataset has 100 images, each with the 1200 $\times$ 1200 resolution.
%
The performance is measured by both bit-rates and distortions. We present the bit-rate in bit-per-pixel (bpp) and distortion in Peak Signal to Noise Ratio (PSNR) and MS-SSIM~\cite{msssim}.

\begin{table}[t]
  \centering\small
   \caption{Computational complexity and BD-rate comparision relative to VVC intra [32] (VTM version 16.2). FLOPs are for 256 $\times$ 256 input resolution.}
  \resizebox{8.5cm}{!} {
  \begin{tabular}{c|ccc}
    \toprule
      Method & Parameters & End-to-end FLOPs & BD-Rate ($\%$) $\downarrow$ \\
      \midrule
      Minnen18~\cite{DavidMinnen2018JointAA} & 25.5M & 29.5B & 10.2\\
      Cheng2020~\cite{ZhengxueCheng2020LearnedIC}  & 29.6M & 60.7B & 2.25\\
      Minnen20~\cite{DavidMinnen2020ChannelwiseAE} & 116M & 38.1B & 0.18\\
      Qian 2021~\cite{YichenQian2021LearningAE} &35.8M & 32.1B & 5.62\\
      Entroformer~\cite{YichenQian2022EntroformerAT} &44.9M & 22.0B & 3.82\\
      InvCompress~\cite{YueqiXie2021EnhancedIE}  & 50M&68.0B & -0.82\\
      Wang 2022~\cite{DezhaoWang2022NeuralDT} & 53M & 523B & -1.22\\
      Oursuformer& 31.7M & 16.6B & -6.32 \\
    \bottomrule
    
  \end{tabular}}
 
  \label{tab:complexityandbdrate}
\end{table}

\begin{table}[t]
  \centering\small
   \caption{The coding time on a 256x256 image on one Nvidia 3090ti GPU.}
  \resizebox{8.5cm}{!} {
  \begin{tabular}{c|ccc}
    \toprule
      Method & Entroformer~\cite{YichenQian2022EntroformerAT} & InvCompress~\cite{YueqiXie2021EnhancedIE}  & Ours\_Uformer \\
      \midrule
      Coding time & 75ms & 1650ms & 59ms\\

    \bottomrule
    
  \end{tabular}}
 
  \label{tab:coding_time}
\end{table}

\subsection{Quantitative Result}

\begin{figure*}[t]
 		\centering
 		\includegraphics[width=0.95\textwidth]{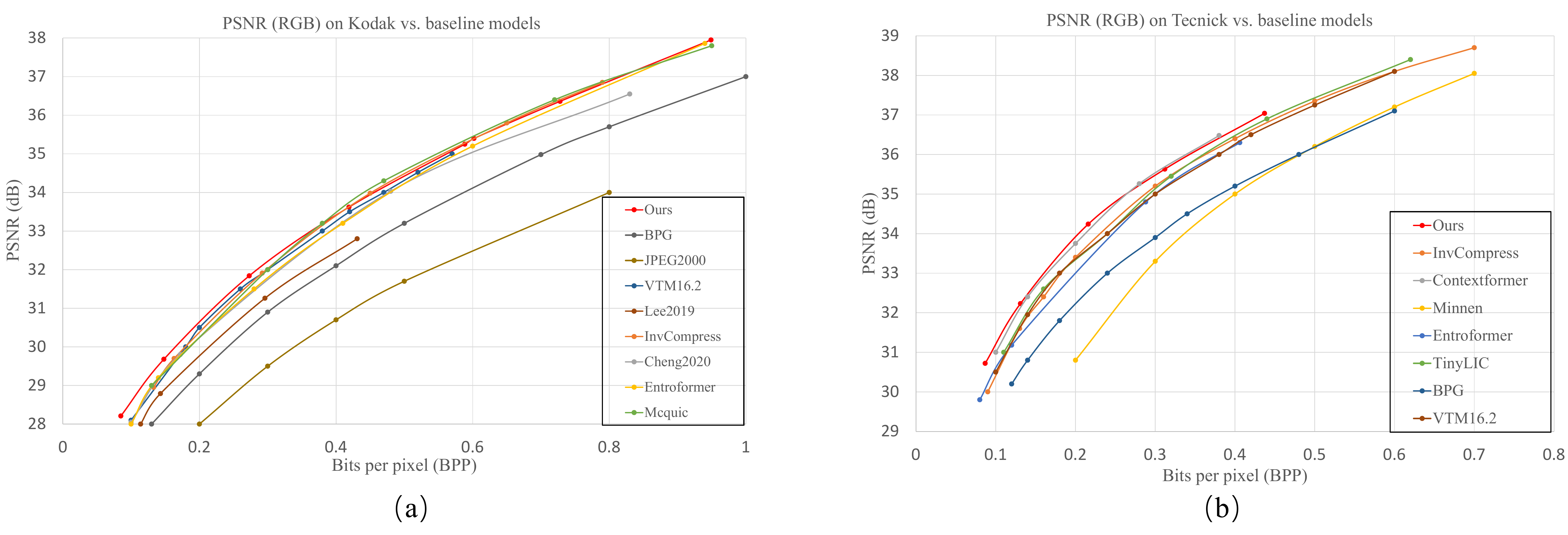}
 		\caption{Comparison of the rate-distortion performance (PSNR-BPP) (a) on Kodak dataset (b) on Tecnick dataset.}
 		\label{fig:rdrate_results}
 \end{figure*}

\begin{figure*}[t]
 		\centering
 		\includegraphics[width=0.95\textwidth]{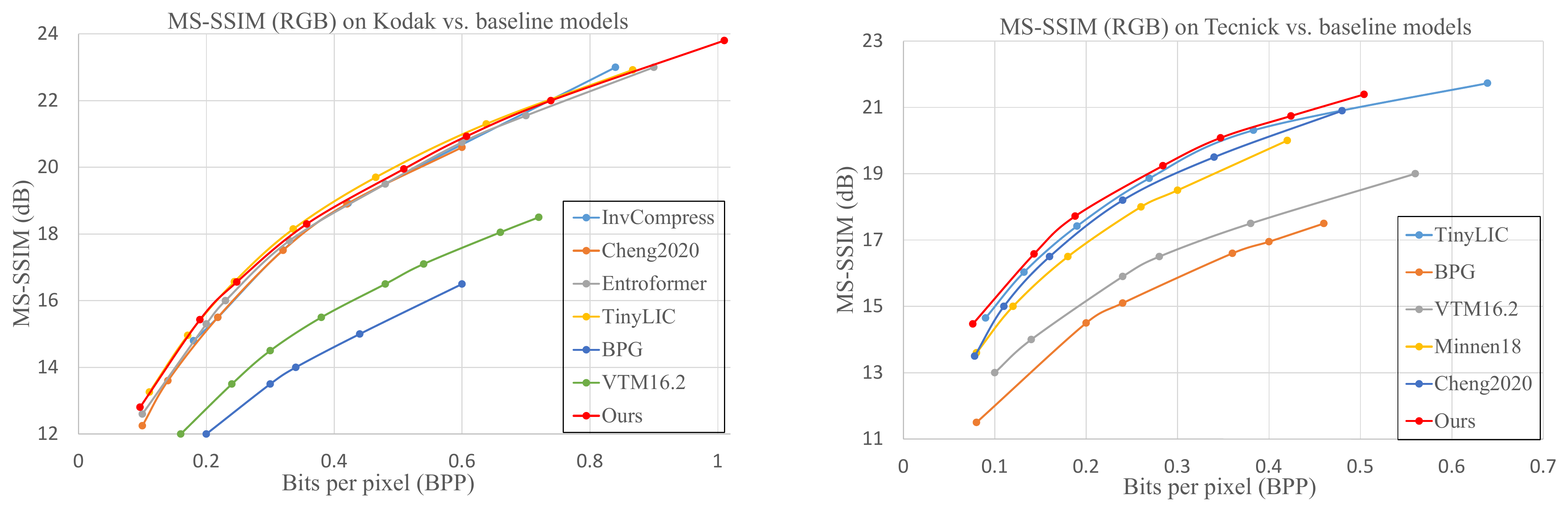}
 		\caption{Comparison of the rate-distortion performance (MS-SSIM-BPP) (a) on Kodak dataset (b) on Tecnick dataset.}
 		\label{fig:rdrate_results_ssim}
 \end{figure*}

\textbf{Rate-Distortion Performance.} We plot the rate-distortion curves (Fig.~\ref{fig:rdrate_results}  and Fig.~\ref{fig:rdrate_results_ssim}) to demonstrate the rate-distortion performance.
The bits-per-pixel (bpp) is used to measure rate. The PSNR and MS-SSIM are used to measure distortion.
We compare with both traditional and state-of-the-art deep learning based methods. The traditional methods are: JPEG 2000~\cite{MajidRabbani2002AnOO}, BPG~\cite{bellard2018},  VVC VTM 16.2~\cite{GaryJSullivan2012OverviewOT}. The deep image compression methods are: Mcquic (CVPR 22)~\cite{XiaosuZhu2022UnifiedMG}, Contextformer (ECCV 22)~\cite{AKoyuncu2022CONTEXTFORMERAT}, Entroformer (ICLR 22)~\cite{YichenQian2022EntroformerAT}, TinyLIC (arXiv 22)~\cite{lu2022high}, InvCompress (ACM MM 21)~\cite{YueqiXie2021EnhancedIE}, Cheng2020 (CVPR 20)~\cite{ZhengxueCheng2020LearnedIC}, Minnen20 (ICIP 20)~\cite{DavidMinnen2020ChannelwiseAE}, Lee (ICLR 19)~\cite{JooyoungLee2019ContextadaptiveEM}, and Minnen18 (NeurIPS 18)~\cite{DavidMinnen2018JointAA}. 
The R-D points are faithfully obtained from either public benchmarks or their original papers.

The results
on Kodak and Tecnick datasets are shown
in Fig.~\ref{fig:rdrate_results}  and Fig.~\ref{fig:rdrate_results_ssim}, respectively.
Fig.~\ref{fig:rdrate_results} shows that our MSE-oriented model is obviously superior to others at a lower bit rate, and comparable to the current best methods at a higher bit rate. %
%
%
Fig.~\ref{fig:rdrate_results_ssim} shows the MS-SSIM-oriented results. On the Tecnick dataset, our method outperforms others. On the Kodak dataset, our method is comparable with the state-of-the-art methods, and outperforms them at low bit rates.
It illustrates that our mask sampling modeling method works well and particularly better at a low bit rate. It is also worth noting that our method requires no special design for the encoder and decoder.
%

\begin{figure}[t]
 		\centering
 		\includegraphics[width=0.46\textwidth]{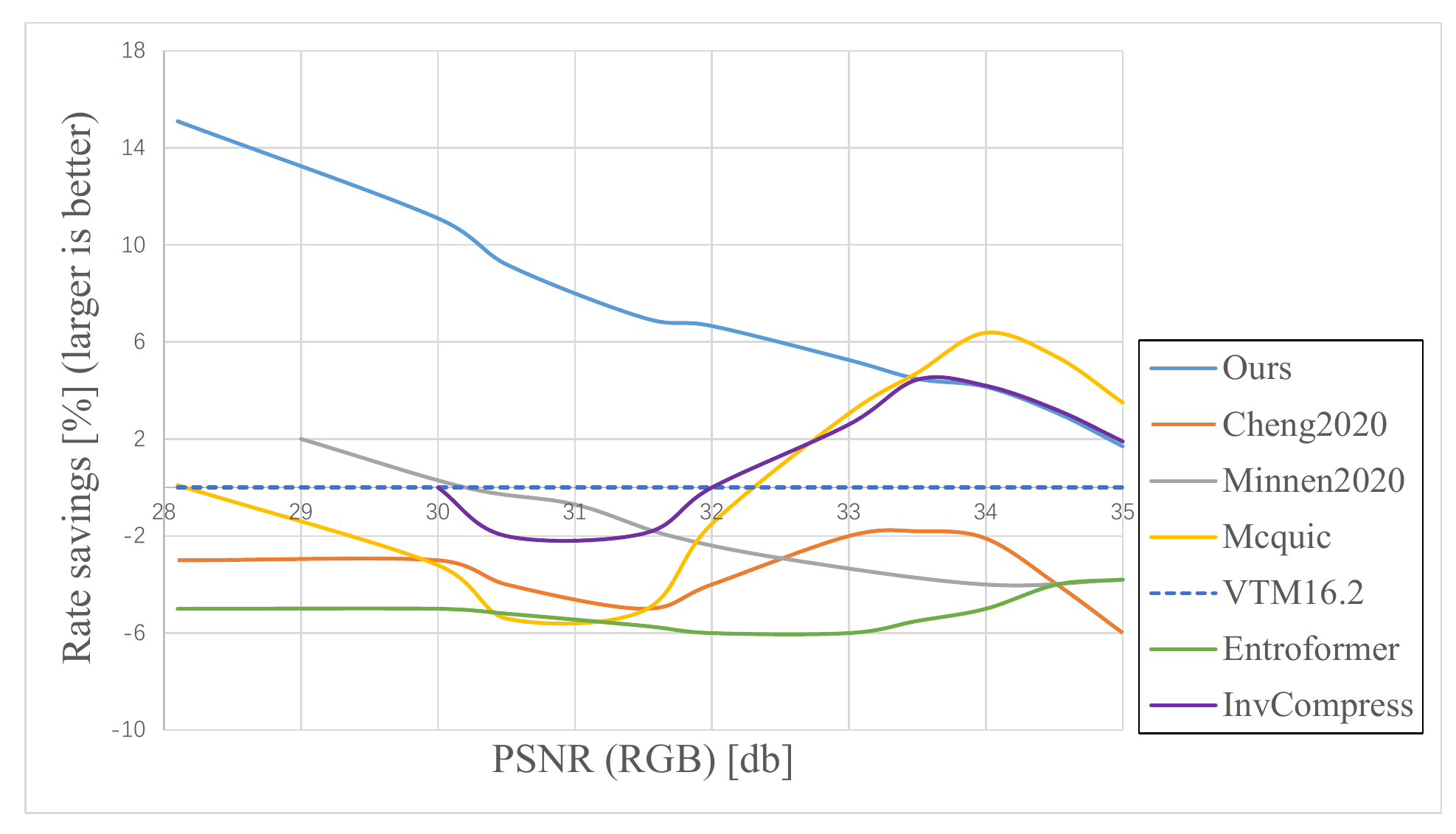}
 		\caption{The rate savings relative to VTM16.2 on the Kodak dataset. Our model reaches about 14\% rate savings for about 28dB and about 2\% rate savings for about 35dB.}
 		\label{fig:ratesaveing_results}
 \end{figure}

\textbf{B-D Rate and Rate Savings.}
We compare the BD-rate~\cite{GisleBjontegaard2001CalculationOA} and computational complexity of our method with state-of-the-art methods in Table~\ref{tab:complexityandbdrate}. In the case where previous methods have multiple model sizes, we choose the highest bit rate models for complexity calculation.
Our method outperforms all others in terms of BD rate on the Kodak dataset.
Among all the methods, only three have negative BD-rate values, and ours outperforms the second-best model Wang 2022~\cite{DezhaoWang2022NeuralDT} by 5.1\%.
The performance in terms of the rate savings relative to VTM16.2 on the Kodak dataset is shown in Fig.~\ref{fig:ratesaveing_results}.
Our model achieves state-of-the-art performance by reaching 14\% rate savings for low bit-rate and
2\% rate savings for high bit-rate over VTM 16.2.

\textbf{Computational Complexity Comparison.}
In Table~\ref{tab:complexityandbdrate}, we compare the model size (`Parameters') and computational complexity (`End-to-end FLOPs') of ours against seven state-of-the-art neural image compression methods.
Our model needs less computation than others.
Though the model sizes of Minnen18 (25.5M) and Cheng2020 (29.6M) are smaller than that of ours (31.7M),  ours has better performance.
We also report the coding time (including encoding time and decoding time) between some methods and ours on one Nvidia 3090ti GPU, as shown in Table~\ref{tab:coding_time}.
Entroformer and InvCompress are implemented reliably using their released official codes.
InvCompress uses a heavy encoder/decoder and cost much time. The coding time of our method (59ms) is less than that of Entroformer (75ms), due to the use of LCMM/LCCM.

 \begin{figure*}[t]
 		\centering
 		\includegraphics[width=0.95\textwidth]{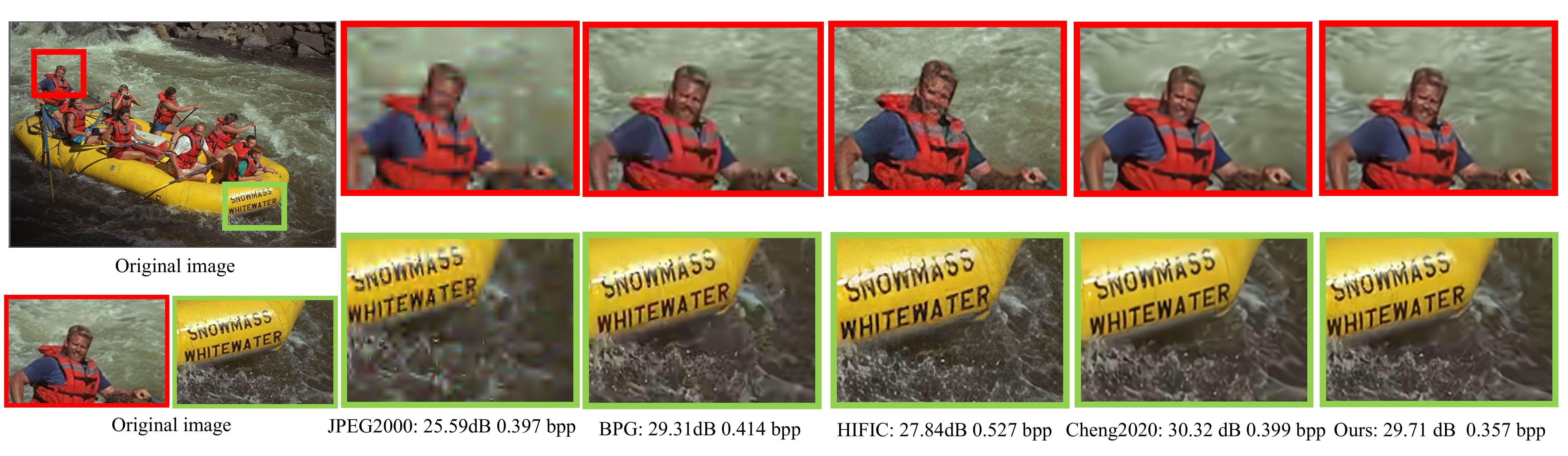}
 		\caption{Visual comparison of reconstructed images from the Kodak dataset. The loss functions of Cheng2020 and ours are MSE losses.}
 		\label{fig:vis_results}
 \end{figure*}

 \begin{figure*}[t]
 		\centering
 		\includegraphics[width=0.92\textwidth]{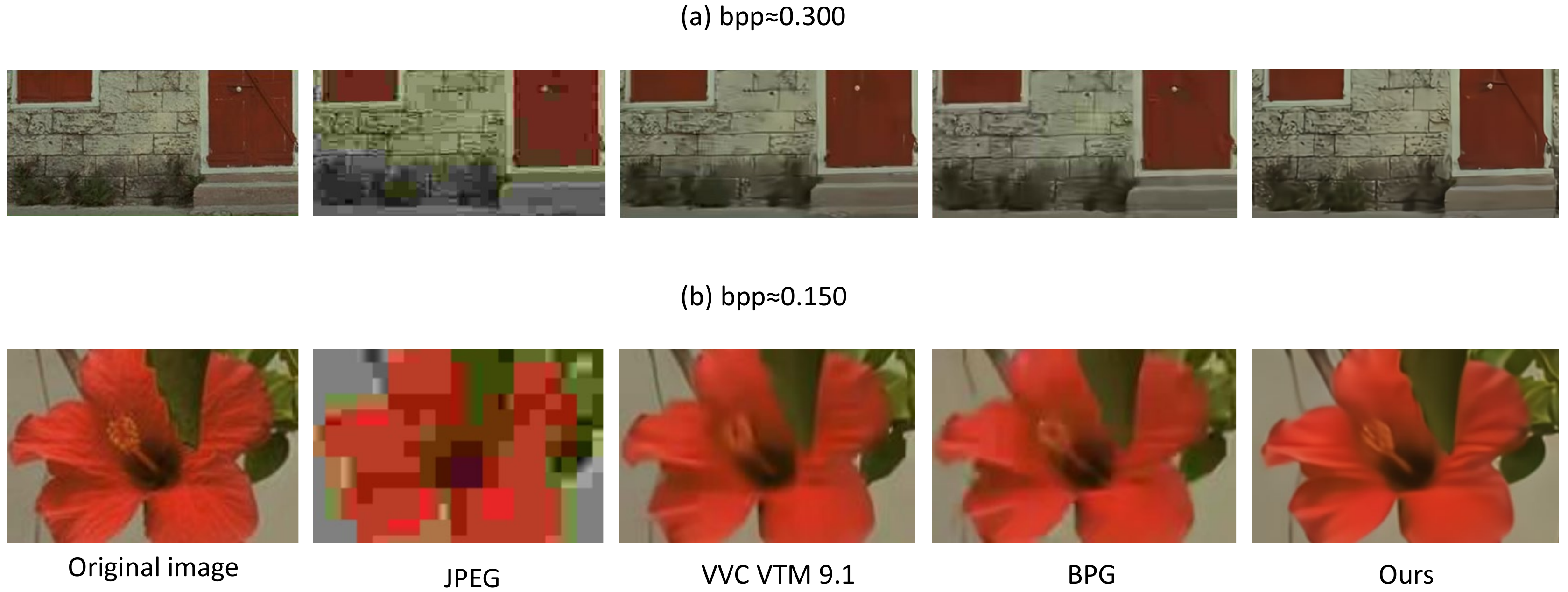}
 		\caption{Visual comparison between the traditional methods and ours from the Kodak dataset.}
 		\label{fig:vis_results2}
 \end{figure*}
 
   \begin{figure*}[t!]
 		\centering
 		\includegraphics[width=0.99\textwidth]{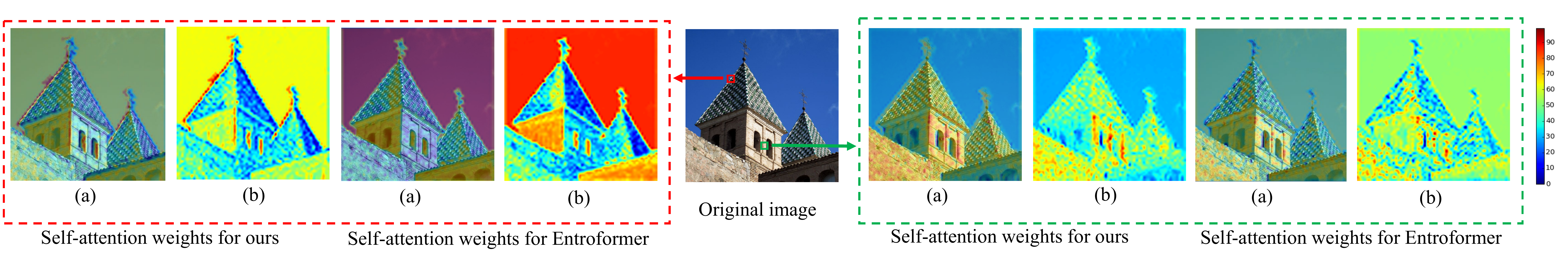}
 		\caption{Visualization of the self-attention weights for two points. (a) the blending of the original image and the weight map; (b) the weight map.}
 		\label{fig:visualization2}
 \end{figure*}

  \begin{figure*}[t!]
 	\centering
 		\includegraphics[width=0.94\textwidth]{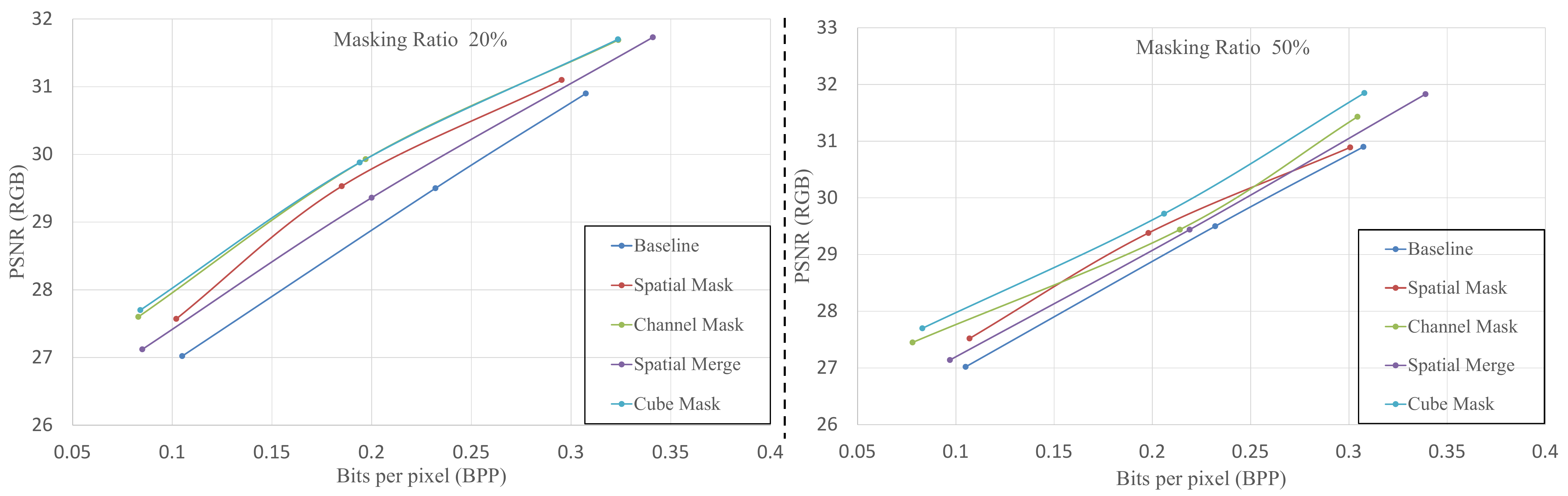}
 		\caption{Ablation study of different mask sampling modeling methods.}
 	\label{fig:maskmode}
 \end{figure*}
 
  \begin{figure}[t]
 		\centering
 		\includegraphics[width=0.46\textwidth]{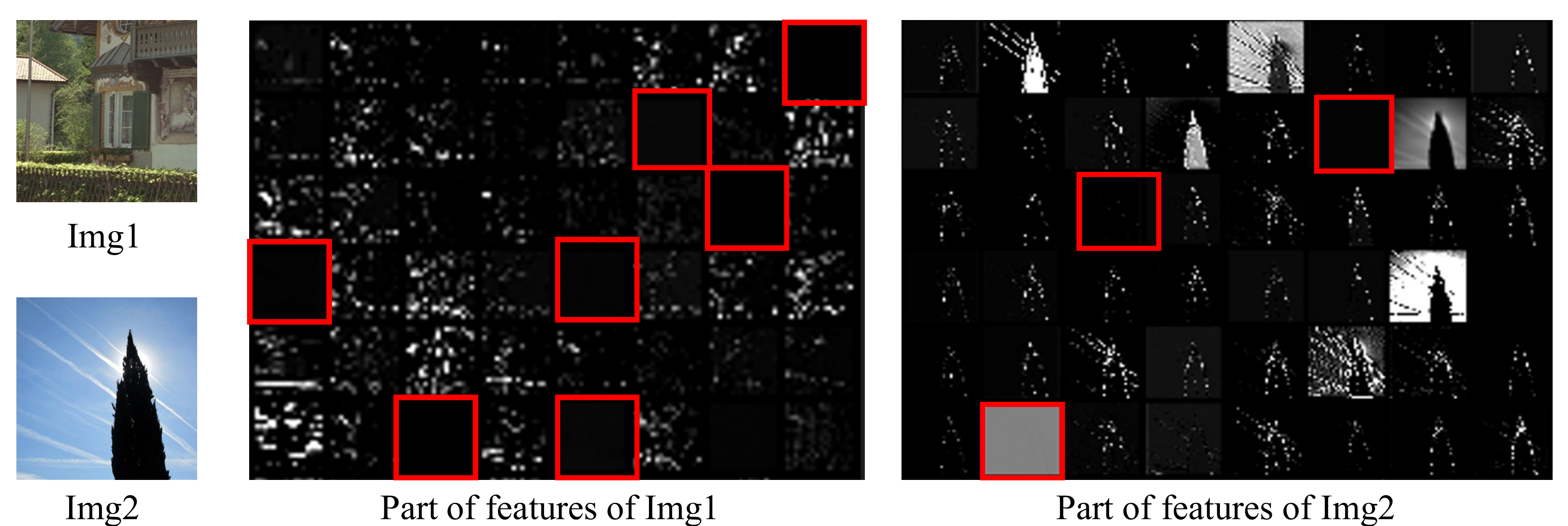}
 		\caption{Visualization of the feature maps after encoders. `Img1' and `Img2' are from Kodak and Tecnick, respectively. The red boxes are abandoned channels in the LCMM module.}
 		\label{fig:visualization1}
 \end{figure}

\subsection{Qualitative Result}
%
In this subsection, we provide qualitative comparisons with state-of-the-art image compression methods: JPEG-2000~\cite{MajidRabbani2002AnOO}, BPG~\cite{bellard2018}, HIFIC~\cite{FabianMentzer2020HighFidelityGI}, and Cheng2020~\cite{ZhengxueCheng2020LearnedIC}.
HIFIC and Cheng2020 are perceptual compression and pixel-based compression methods, respectively.
The results are shown in Fig. ~\ref{fig:vis_results}. Results of `JPEG2' and `BPG' contain blurs. There exist noises on the face of the prediction of `HIFIC'. Our result does not have any of these issues and can well preserve image details. 
%
%
Cheng2020 has a comparable result with ours, but ours has a much smaller bpp. 
In Fig.~\ref{fig:vis_results2} , we provide qualitative comparisons with traditional image compression methods: JPEG~\cite{wallace1992jpeg}, BPG~\cite{bellard2018}, and VVC VTM 9.1~\cite{liu2019vvc}.
Our approach exhibits superior performance and preserves more details.

\subsection{Ablation Study}
We conduct ablation studies to investigate each module's impact. 
In this section, due to the significant time cost required to train on the complete Imagenet-val dataset, we use the ImageNet-Val-8k dataset (following \cite{He_2021_CVPR}, a subset of ImageNet-Val) for training and fine-tuning.

\textbf{Mask Sampling Modeling Methods.}
In order to compare the performance of different mask sampling modeling methods of the CMSM module, we conducted pre-training and finetuning under masking ratio of 50\% and 20\%, as shown in Fig.~\ref{fig:maskmode}. During these experiments, only the CMSM in Fig. 2(b) is replaced with other mask sampling modeling Methods while keeping other modules unchanged.
For pre-training, we used a lambda value of 0.3 and the MSE loss function. During fine-tuning, we used lambda values of 0.001, 0.003, and 0.006. The results show that the Cube Mask method has the best performance under masking ratio of both 50\% and 20\%.
This masking method allows masked elements to obtain information from both other channels and spatial locations simultaneously.

\begin{table}[t]
    \caption{Ablation of cube masking ratio of the CMSM module. All models are optimized with $\lambda$ = 0.01 (PNSR $\approx$ 33.5 on Kodak).}
  
  \centering
  \resizebox{8.1cm}{!} {
  \begin{tabular}{c|cc|cc}
    \toprule
      \multirow{2}{*}{Masking Ratio}& \multicolumn{2}{c|}{Dim=320} & \multicolumn{2}{c}{Dim=364} \\
      &bpp~$\downarrow$ & $\Delta$ bpp (\%)~$\downarrow$& bpp~$\downarrow$ & $\Delta$ bpp (\%)~$\downarrow$ \\
    \midrule
      70\%&0.4343 & 1.04   & 0.4664 & 6.43\\
      50\%&0.4133 & -3.83  & 0.4141 & -5.49\\
      20\%&0.4149 & -3.46  & 0.4142 & -5.47\\
      10\%&0.4177 & -2.81  & 0.4196 & -4.24\\
      0\% (baseline)&0.4298 & 0.0 &0.4382 & 0.0\\
    \bottomrule
  \end{tabular}}
  
  \label{tab:maskratio}

\end{table}

\textbf{Masking Ratios.}
In order to explore the impact of different masking ratios on the performance of the CMSM module, we tried five different masking ratios, including 70\%, 50\%, 20\%, 10\%, and 0\%.
We conducted experiments under $\lambda$ equal to 0.01. Because the PSNR results for different masking ratios are very close (PSNRs are about 33.5dB, with a deviation of less than 0.3\%), we compare the reduction of BPP.
From Table~\ref{tab:maskratio}, we can see that a high masking ratio (70\%) results in performance degradation, with an increase of about 1\% and 6.4\% in BPP for channel numbers 320 and 364, respectively.
Compared to other masking ratios, the performance of the 50\% masking ratio is the best, with a decrease of 3.83\% and 5.49\% in BPP for channel numbers of 320 and 364, respectively.

\begin{figure}[t]
 		\centering
 		\includegraphics[width=0.45\textwidth]{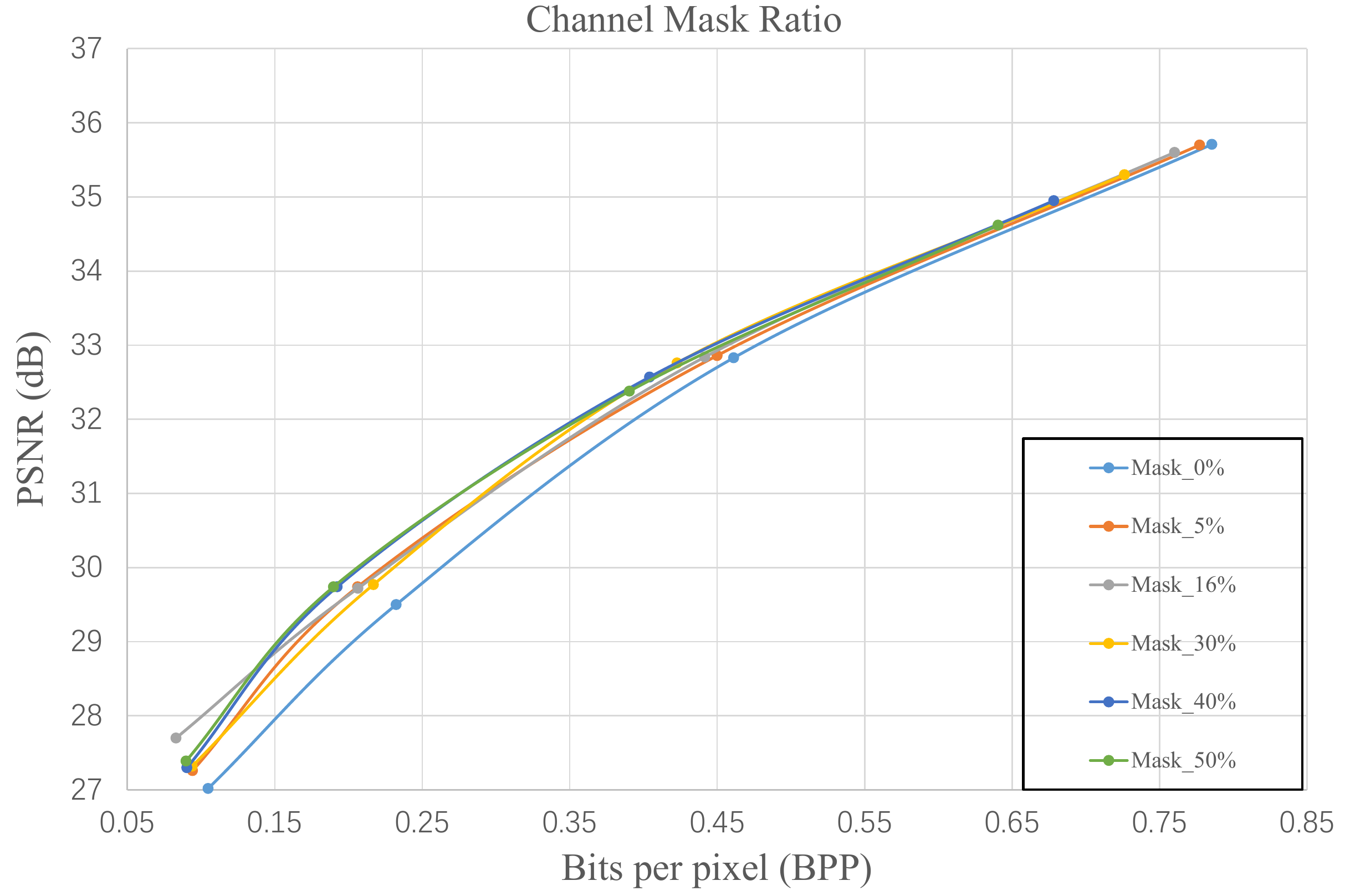}
 		\caption{Ablation study of different channel mask ratio of the LCMM and LCCM modules.}
 		\label{fig:channelmaskratio}
 \end{figure}

We also explore the suitable masking ratios of the LCMM and LCCM modules.
In Figure ~\ref{fig:channelmaskratio}, we investigate the optimal masking ratios for the LCMM and LCCM modules. The `Mask\_xx\%' represents the percentage of channels that require masking. At a bpp of approximately 0.1, a masking ratio of 16\% yields the best performance, while a masking ratio of 50\% performs best at a bpp of around 0.2. As the bpp increases, the performance of the 30\% and 40\% masking ratios also improves, approaching the performance of the 50\% masking ratio. Furthermore, the performance of networks with different masking ratios gradually approaches the baseline performance at high bpp, indicating that our method performs better at low bpp than at high bpp.
%

 \begin{figure}[t]
 		\centering
 		\includegraphics[width=0.45\textwidth]{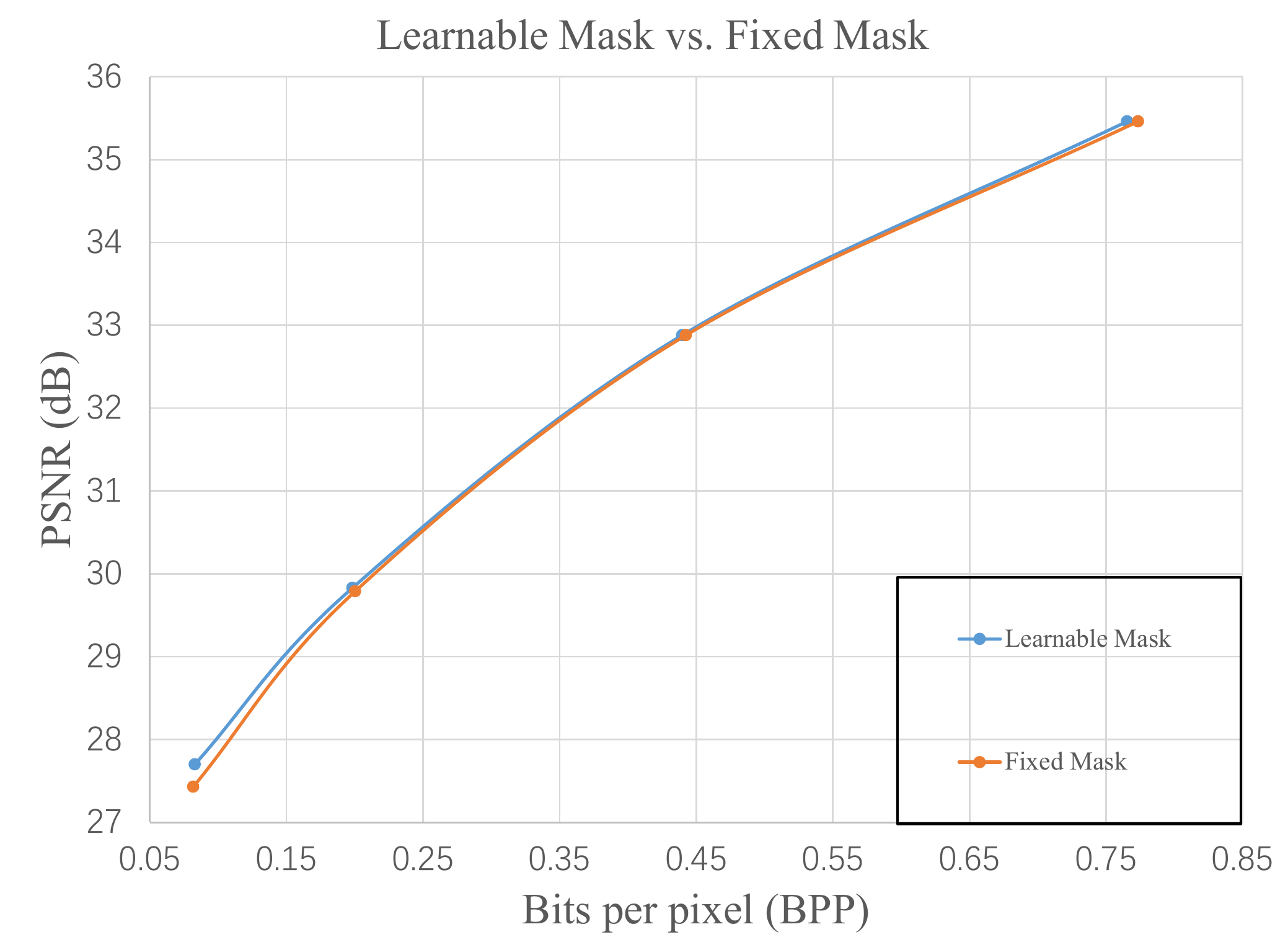}
 		\caption{Ablation study of learnable mask and fixed mask.}
 		\label{fig:tokenmask}
 \end{figure}

\textbf{Learnable channel mask vs. fixed channel mask.}
We also conduct ablation studies to explore whether the learned or randomized vectors in the LCMM and LCCM modules are better.
In the randomized mask, the vector $v_{mat}$ is randomized at the initial stage and fixed for training, finetuning, and testing. The experimental results show that the performance of the learned vector is slightly better. More detailed results can be found in the supplementary materials.

\textbf{Removing CMSM or
LCMM/LCCM.}
The 0\% (baseline) in Table \ref{tab:maskratio} is equivalent to removing the CMSM module, and the blue curve (Mask\_0\%) in Fig. 14 corresponds to removing both LCMM and LCCM. These two experiments show the effectiveness of our full model.

\subsection{Visualization}

\textbf{Visualization of Self-attention Weights.} In Fig.~\ref{fig:visualization2}, we visualize the first self-attention maps of the hyper encoder of ours and Entroformer~\cite{YichenQian2022EntroformerAT}.
These maps show how the model searches for similar context for the current latent elements.
It shows that ours performs better than Entroformer to find more accurate structure or color information. 

\textbf{Visualization of the Channel Mask.} In Fig.~\ref{fig:visualization1}, we visualize the feature maps generated by the encoder to show how the LCMM mask channels. The red boxes are abandoned channels in the LCMM module.
We can see that most of the discarded feature maps have almost no content, while a few contain image-independent content (such as the feature map highlighted by the last red box of `Img2').


\section{Conclusions}
In this paper, we explore effective mask sampling modeling for neural image compression.
First, we propose a novel Cube Mask Sampling Module (CMSM) that applies mask sampling modeling to image compression. 
Additionally, we propose the Learnable Channel Mask Module (LCMM) and
the Learnable Channel Completion Module (LCCM) to reduce channel redundancy.
Experiments illustrate that our method achieves competitive performance with lower computational complexity compared to state-of-the-art
methods.
Future work includes exploring more effective masking methods and testing our proposed modules on more architectures and low-level vision tasks.

\bibliographystyle{IEEEtran}
\bibliography{egbib}

\vfill

\end{document}